\documentclass[10pt,twocolumn,letterpaper]{article}
\makeatletter
\providecommand{\@LN}[2]{}
\makeatother
\usepackage{cvpr}              

\definecolor{cvprblue}{rgb}{0.21,0.49,0.74}
\usepackage[pagebackref,breaklinks,colorlinks,allcolors=cvprblue]{hyperref}
\usepackage{multirow}
\usepackage{graphicx} 
\usepackage{algpseudocode}
\usepackage{amsmath} 
\usepackage{hyperref}       
\usepackage{url}            
\usepackage{booktabs}       
\usepackage{amsfonts}       
\usepackage{nicefrac}       
\usepackage[ruled,vlined]{algorithm2e}
\usepackage{amsmath}
\usepackage{mdframed}
\usepackage{amsmath, amssymb, bm, mathrsfs}
\usepackage{caption}
\usepackage{subcaption}
\newcommand{\vect}[1]{\bm{#1}}  
\newcommand{\gX}{\mathscr{X}}   
\newcommand{\gY}{\mathscr{Y}}   
\def\paperID{10584} 
\def\confName{CVPR}
\def\confYear{2025}

\title{Prompt2Perturb (P2P): Text-Guided Diffusion-Based Adversarial Attacks on Breast Ultrasound Images}

\author{
    Yasamin Medghalchi$^{1}$ \quad Moein Heidari$^{1}$ \quad Clayton Allard$^{1}$ \quad Leonid Sigal$^{1,2}$ \quad Ilker Hacihaliloglu$^{1}$\\
    $^1$University of British Columbia, Vancouver, BC, Canada \\
    $^2$Vector Institute for AI, Toronto, ON, Canada \\
    {\tt\small yasimed@student.ubc.ca}\\
    {\tt\small \{moein.heidari, ilker.hacihaliloglu\}@ubc.ca}\\
    {\tt\small clayton.allard@stat.ubc.ca, lsigal@cs.ubc.ca}
}

\begin{document}
\maketitle
\begin{abstract}


Deep neural networks (DNNs)  offer significant promise for improving breast cancer diagnosis in medical imaging. However, these models are highly susceptible to adversarial attacks—small, imperceptible changes that can mislead classifiers—raising critical concerns about their reliability and security. Traditional attacks rely on fixed-norm perturbations, misaligning with human perception. In contrast, diffusion-based attacks require pre-trained models, demanding substantial data when these models are unavailable, limiting practical use in data-scarce scenarios. In medical imaging, however, this is often unfeasible due to the limited availability of datasets. Building on recent advancements in learnable prompts, we propose Prompt2Perturb (P2P), a novel language-guided attack method capable of generating meaningful attack examples driven by text instructions. During the prompt learning phase, our approach leverages learnable prompts within the text encoder to create subtle, yet impactful, perturbations that remain imperceptible while guiding the model towards targeted outcomes.
In contrast to current prompt learning-based approaches, our P2P stands out by directly updating text embeddings, avoiding the need for retraining diffusion models. Further, we leverage the finding that optimizing only the early reverse diffusion steps boosts efficiency while ensuring that the generated adversarial examples incorporate subtle noise, thus preserving ultrasound image quality without introducing noticeable artifacts. We show that our method outperforms state-of-the-art attack techniques across three breast ultrasound datasets in FID and LPIPS. Moreover, the generated images are both more natural in appearance and more effective compared to existing adversarial attacks.
Our code will be publicly available at~\href{https://github.com/yasamin-med/P2P}{\textcolor{magenta}{GitHub}}.
\end{abstract}    
\section{Introduction}
\label{sec:intro}
\begin{figure}
    \centering
    \includegraphics[width=1\linewidth]{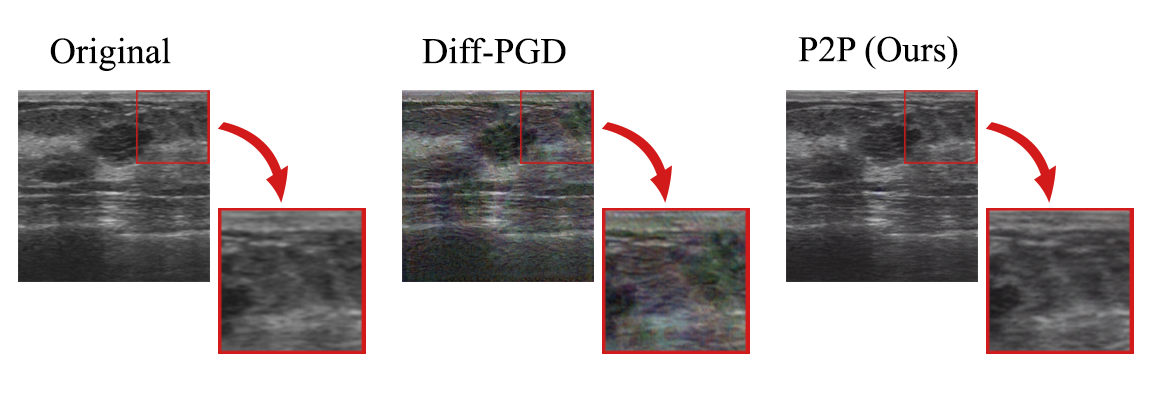}
    \caption{Illustration of P2P in an adversarial attack against Diff-PGD; note there is no exhibited change of image semantics in our method. }
    \label{fig:enter-label}
\end{figure}
Mammography remains the primary tool for breast cancer screening, offering widespread accessibility and proven effectiveness in detecting early-stage tumors \cite{elhatw2023advanced}. However, it has some limitations, including patient discomfort and reduced diagnostic accuracy in individuals with dense breast tissue \cite{lee2016digital}. Ultrasound has become a useful supplementary technique, as it provides better patient comfort and avoids the risks associated with radiation exposure. Its ability to capture real-time images makes it beneficial for further investigating abnormalities observed on mammograms \cite{elhatw2023advanced}. Nonetheless, the performance of ultrasound can be inconsistent, with sensitivity and specificity affected by the operator’s expertise, device configuration, and changes in tissue characteristics due to breast density. Furthermore, variability in interpretation among different radiologists poses additional challenges \cite{svensson2010advanced}. In response to these issues, computational approaches, especially deep learning models, are gaining traction as promising tools to improve diagnostic outcomes and address these existing gaps \cite{svensson2010advanced, cheng2016computer, kazerouni2023diffusion}.\\
Ensuring the security and robustness of medical deep neural networks (DNNs) is a key concern in healthcare. Recent research in medical imaging tasks has revealed that even state-of-the-art DNNs are highly susceptible to adversarial attacks \cite{ma2021understanding}. Specifically, medical imaging DNNs are more susceptible to adversarial threats than those designed for natural images \cite{ma2021understanding}. In addition, the limited availability of publicly accessible medical images, which poses a challenge for effective training, has pushed modern algorithms to rely on pre-trained models based on large datasets of natural images—a technique known as transfer learning. However, due to the substantial difference between natural and medical images, deep learning models developed through transfer learning tend to be more vulnerable to adversarial attacks \cite{chen2024adversarial}.\\
To address this vulnerability, various adversarial attack methods have been developed.
Traditional gradient-based methods such as fast gradient sign method (FGSM) \cite{goodfellow2014explaining}, Carlini-wagner \cite{carlini2017towards}, and projected gradient descent (PGD) \cite{madry2017towards} perturb
original images within predefined perturbation bounds. Although they can be easily generated using gradient-based techniques, these attacks often deviate significantly from the true data distribution of natural images, leading to a compromise between their effectiveness and subtlety \cite{xue2024diffusion}.
Generative models such as generative adversarial networks (GANs) \cite{goodfellow2014generative} have been used to generate adversarial examples, based on their ability to produce high-quality images, by perturbing latent space \cite{song2018constructing,zhao2017generating,ashrafian2024vision}. However, modifying latent codes changes the high-level semantics of generated images in a way that is easily noticeable to human observers \cite{karras2019style}.
Recently, diffusion-based models have been employed in the literature to generate realistic adversarial images \cite{chen2023advdiffuser,xue2024diffusion,lin2023sd}. In models like AdvDiffuser \cite{chen2023advdiffuser} and Diff-PGD \cite{xue2024diffusion}, the PGD method is incorporated into the diffusion process to enhance the realism of the generated images. Furthermore, approaches such as Instruct2Attack \cite{liu2023instruct2attack} leverage latent diffusion models \cite{rombach2022high} to deceive the target model by optimizing over input prompts. These text-guided attacks aim to produce adversarial semantic image edits, such as altering the color of a truck in an image. While image editing holds significant value in natural image domains, it is less applicable in the medical field, as altering the semantics of medical images can result in unrealistic representations. Additionally, these models are often not well-adapted to clinical terminology, making such modifications less meaningful in medical contexts \cite{wang2022medclip}.\\
These limitations can be addressed by reversing the diffusion model to obtain context-relevant text tokens, or prompts, specifically designed for the medical domain to generate adversarial images. Optimizing these prompts presents a challenge due to the intricate structure of the generative model, which includes a two-part pipeline: a conditioning network (the CLIP text encoder \cite{radford2021learning}) and a generative unit (U-Net \cite{ronneberger2015u}), both of which complicate gradient flow to the input layer. To tackle these obstacles, we freeze the generative model and make the following contributions:

\begin{itemize}
    \item Our method requires no fine-tuning and efficiently generates attack images by directly updating text embeddings, unlike previous diffusion-based approaches reliant on large pre-trained models on the same domain. Additionally, we find that optimizing only the initial reverse diffusion steps is sufficient, improving efficiency and maintaining image quality without the need for all reverse diffusion phases.
    
    \item Based on our findings, we proposed Prompt2Perturb (P2P) to create challenging adversarial examples optimized for clinical terminology that are particularly challenging to distinguish as attacked, thus highlighting limitations in current adversarial attack mechanisms.
    
    \item We demonstrate empirically that our method produces semantically significant attack images that are perceptually similar but varied versions of a target image (see \cref{fig:enter-label}). By incorporating relevant clinical terminology into the text embedding of Stable Diffusion, we ensure that the generated images retain clinical accuracy and realism, thereby enhancing the effectiveness of the attack in a medical context.
    
\end{itemize}

\section{Related Work}
\label{sec:relatedwork}

\begin{figure*}[t]
\begin{center}
\includegraphics[width=0.9\linewidth]{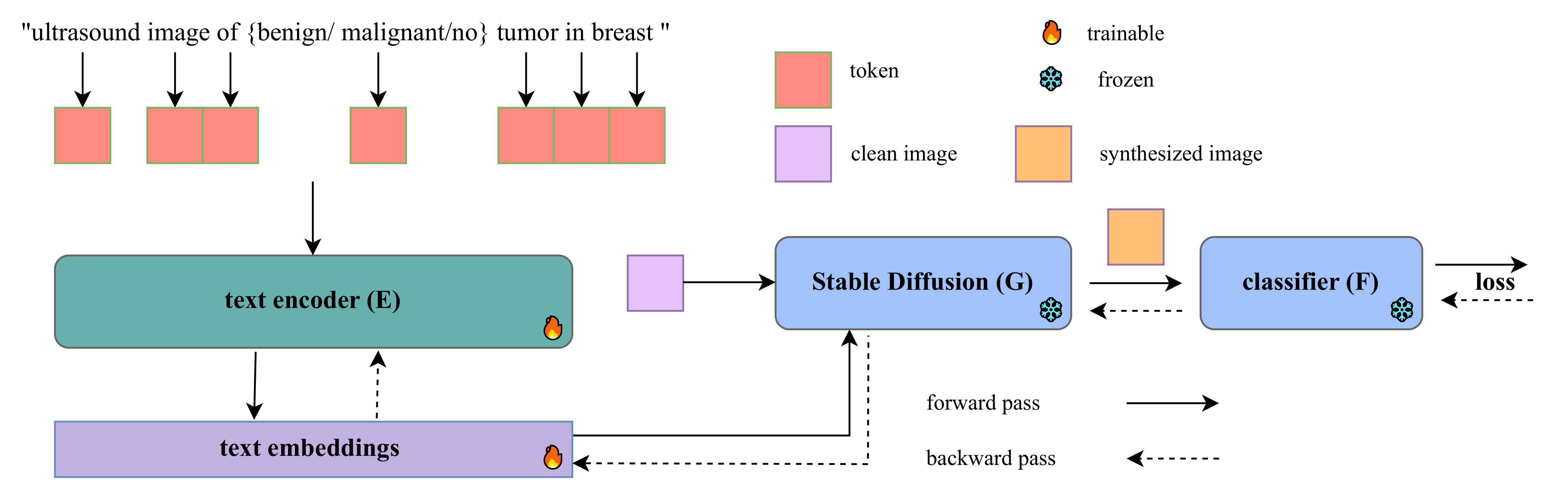}
\end{center}
\vspace{-.6cm}
   \caption{Overall framework of the proposed method. Image adapted from \cite{lin2023sd}}
   \vspace{-.5cm}
\label{fig:architecture}
\end{figure*}

\textbf{Adversarial attack.}
DNN models are widely recognized as being susceptible to adversarial attacks \cite{ren2020adversarial}. These attacks can manipulate the model's predictions by introducing subtle modifications that are imperceptible to human observers.
Adversarial attacks pose a major barrier to the widespread adoption of deep learning models in real-world applications, and considerable research has been dedicated to addressing this ongoing challenge. Adversarial attacks can be broadly classified into black-box and white-box types. White-box attacks, such as limited-memory
Broyden–Fletcher–Goldfarb–Shanno (L-BFGS) \cite{szegedy2013intriguing}, FGSM \cite{goodfellow2014explaining}, PGD \cite{madry2017towards}, C-W \cite{carlini2017towards}, and DeepFool \cite{moosavi2016deepfool}, rely on full access to the target model's architecture and parameters, making them gradient-based. In contrast, black-box attacks do not require this level of access and generate adversarial samples through query-based methods. Despite these differences, many white-box attacks are still effective in black-box settings due to the transferability of adversarial examples across models. Adversarial training, which involves training a model using adversarially generated examples, is one of the limited methods available to defend against adversarial attacks \cite{bai2021recent}.
In medical domain, several approaches have been developed to generate adversarial examples targeting computer-aided diagnosis models which can be consistently taxonomized with that of adversarial attacks \cite{dong2023adversarial}.  Specifically, Zhou et al. \cite{zhou2021machine} introduced two GAN-based models operating at different resolutions, capable of generating highly realistic adversarial images that can mislead breast cancer diagnosis. MIRST \cite{sun2022mirst} was proposed to defend untargeted adversarial attacks on breast ultrasound images. Lastly, Hao et al. \cite{hao2024adversarially} introduce an adversarially robust feature learning method where a feature correlation metric is included as an objective function to promote the learning of robust features while minimizing the influence of spurious ones for the application of breast cancer diagnosis.
\\
\textbf{Adversarial attack with diffusion models.}
The success of diffusion models has motivated their exploration in adversarial deep learning, demonstrating potential for both attacking \cite{zhuang2023pilot,xue2023diffusion} and defending \cite{wang2023better,wu2023defending} against adversarial examples.
These methods are broadly used to enhance the imperceptibility of unconstrained adversarial perturbations \cite{chen2024diffusion}, transfer the style
from the reference image into the original image to create natural-looking adversarial images \cite{wang2023advst}, or attack the diffusion-based purification method \cite{kang2024diffattack}.
\\
\textbf{Prompt learning.}
Recent progress in controllable image generation and editing using diffusion models \cite{ho2020denoising,sohl2015deep} has created new possibilities for developing semantic adversarial attacks. Prompt learning was initially introduced as an alternative to full fine-tuning and linear probing, aiming to leverage pre-trained language models in natural language processing (NLP) \cite{lester2021power}. Recently, prompt learning techniques have been adapted for adversarial purposes. For instance, BadCLIP \cite{bai2024badclip} targets the CLIP \cite{radford2021learning} model by injecting a backdoor during the prompt learning phase. In the medical field, BAPLe \cite{hanif2024baple} utilizes learnable prompts within the text encoder and introduces imperceptible learnable noise triggers into input images, effectively leveraging the full capabilities of medical foundation models. Additionally, PromptSmooth \cite{hussein2024promptsmooth} uses prompt learning to efficiently achieve certified robustness in medical vision-language models by leveraging semantic concepts.

\section{Methodology}
\label{sec:method}

\subsection{Preliminary}

We explore the challenge of image classification. Consider an image sample $\vect{x} \in \gX := \mathbb{R}^{H \times W \times C}$, paired with its label $y \in \gY := \{1, \dots, |\gY|\}$, where $H$, $W$, and $C$ represent the image’s height, width, and number of channels, respectively. Let $f: \gX \rightarrow \gY$ denote the image classifier. Given a clean input image $\vect{x}$ and its true label ${y}$, the attacker’s objective is to generate an adversarial version $\vect{x}_\text{adv}$ of $\vect{x}$ that misleads $f$, so that $f(\vect{x}_{\text{adv}}) \neq {y}$. Traditional approaches either necessitate adding pixel-level adversarial noise $\delta$~\cite{goodfellow2014explaining, carlini2017towards} to the input image $\vect{x}$, resulting in $\vect{x}_\text{adv} = \vect{x} + \delta$, or manipulating data to embed subtle biases or vulnerabilities.
In this work, we instead focus on prompt-guided semantic attacks, aiming to create meaningful, interpretable perturbed images guided by language descriptions.

\subsection{Prompt2Perturb (P2P)}

\Cref{fig:architecture} provides a high-level overview of the proposed approch. Our approach seeks to generate an adversarial image, $\vect{x}_\text{adv}$, such that the modification from the original image $\vect{x}$ is guided by a text-based embedding instruction $\vect{C}$. The core idea of Prompt2Perturb (P2P) is to optimize these text embeddings specifically for the modified input to steer the classifier toward an incorrect prediction. In the subsequent sections, we will provide a detailed breakdown of each component of our proposed method.
\\
\textbf{Latent Diffusion Models.}
We utilize the learned latent space of a generative model, defined as $\mathcal{Z}={\vect{z} \ | \ \vect{z}=\mathcal{E}(\vect{x}), \vect{x} \in \mathcal{X}} \subseteq \mathbb{R}^{h \times w \times c}$, where $\mathcal{E}$ represents the image encoder and $h \times w \times c$ denotes the dimensions of the latent space. Unlike the image space, this latent space offers a compact, lower-dimensional representation that emphasizes the critical semantic details, making it ideal for controlled semantic alterations. These modifications are enabled through a pretrained conditional latent diffusion model (LDM)~\cite{brooks2023instructpix2pix, rombach2022high}, which facilitates meaningful and guided editing in the latent domain.

The encoder is regularized via either a KL-divergence penalty or vector quantization, as suggested by prior works \cite{van2017neural}. The corresponding decoder, denoted as $\mathcal{D}$, learns to reconstruct input images from the latent representations, ensuring $\mathcal{D}(\mathcal{E}(x)) \approx x$, for high fidelity reconstruction. Subsequently, a diffusion model is trained to generate latent codes that reside within the learned latent space. This diffusion model is flexible and can be conditioned using auxiliary information, such as class labels, segmentation masks, or features from a co-trained text-embedding model. Specifically, let $c_\theta(p)$ represent a network mapping conditioning input $p$ to its respective conditioning vector. The objective function for the Latent Diffusion Model (LDM) can thus be formulated as:

\[
\mathcal{L}_{\text{LDM}} := \mathbb{E}_{z \sim \mathcal{E}(x), p, \epsilon \sim \mathcal{N}(0,1), t} \left[ \left\lVert \epsilon - \epsilon_\theta(z_t, t, c_\theta(p)) \right\rVert_2^2 \right],
\]

where $t$ is the time step, $z_t$ is the latent noised to time $t$, $\epsilon$ is the unscaled noise sample, and $\epsilon_\theta$ is the denoising network. Intuitively, the objective here is to correctly remove the noise added to a latent representation of an image. While training, $c_\theta$ and $\epsilon_\theta$ are jointly optimized to minimize the LDM loss. At inference time, a random noise tensor is sampled and iteratively denoised to produce a new image latent, $z_0$. Finally, this latent code is transformed into an image through the pre-trained decoder $x' = \mathcal{D}(z_0)$.
Previous diffusion-based attack methods leverage extensively pre-trained models on natural images, incorporating specialized attack strategies and employing a reverse diffusion process to generate adversarial examples that resemble the underlying data distribution. However, this approach is impractical in the medical domain due to the lack of extensively trained models caused by inherent data scarcity and the presence of diverse imaging modalities. Unlike these previous approaches, our method is well-suited to handle this diversity without requiring retraining of a diffusion model or relying on large-scale pre-trained models. This robustness stems from our approach, which involves updating the text embeddings directly, rather than merely adding noise to the latent space.
We next review the early stages of such a text encoder, and our choice of attack space.
\\
\textbf{Text optimization.}
The text encoder processes the input prompt $\mathbf{P}$, which is fed into the latent diffusion model. A text tokenizer transforms the prompt strings into tokens, representing indices in a predefined vocabulary of size $|V|$. Each token is then mapped to its corresponding embedding vector $\mathbf{c}_i \in \mathbf{C}$, where $i \in {1, \ldots, |V|}$ within the embedding space $\mathcal{C}$. Previous work on textual inversion in the embedding space $\mathcal{C}$~\cite{gal2022image,voynov2023p+} attempts to learn new concepts for a set of images by associating placeholder strings with newly learned embedding vectors. This effectively expands the existing vocabulary with additional concepts. However, these newly learned concepts have been found to be uninformative in the medical domain, as the learned embeddings are often distant from the embeddings corresponding to the original vocabulary. In contrast, our approach aims to optimize the embedding vectors ${\mathbf{c}}_i$, $i \in {1, \ldots, L}$, from the existing vocabulary for a prompt of maximum length $L$ specific to the medical domain. This adaptation addresses the limitations of existing diffusion model vocabularies that lack specific medical terms (e.g., the word ``benign" is not present in the vocabulary of the Stable Diffusion text encoder, despite its frequent use in medical contexts). Therefore, our method ensures applicability across diverse imaging modalities and scenarios within the medical domain, as it remains independent of the specific encoder choice and is instead reliant on the latent diffusion backbone used for prompt optimization.
\\
\textbf{Minimal reversal steps.}
Previous studies \cite{mahajan2024prompting} have demonstrated that the diffusion model's sensitivity to conditioning intensifies significantly during the later stages of the diffusion process, particularly when noise levels are higher. Conversely, in the initial timesteps, the diffusion loss remains relatively stable across various prompts applied to the same image. Inspired by this observation, we hypothesize that not all reverse diffusion steps are necessary for generating an adversarial image. Optimizing only for the early timesteps of denoising process (akin to later timesteps of diffusion process) has two-fold advantage. First, given that we are optimizing the early layers of a very deep neural network, the resulting gradients tend to be quite small which contributes to our method's efficiency compared to others that require training the entire diffusion model. Additionally, early timesteps (the start of the denoising process) are associated with low-frequency components, focusing on reconstructing coarse shapes and structures \cite{qian2024boosting}. As the process moves into the later timesteps, the model increasingly refines high-frequency details. By introducing adversarial perturbations during this stage, our strategy generates examples that effectively mimic common scenarios in ultrasound imaging, where acquisition settings—such as the transducer's operating frequency and its orientation relative to the anatomy—significantly influence the high-frequency content captured in the images \cite{lieu2010ultrasound}.
We further substantiate this reasoning in the subsequent sections.

\begin{algorithm}[h]
\caption{Text-Guided Diffusion-Based Attack Pipeline}
\KwIn{Attacked classifier $f$, Image $x$, Latent $z$, Prompt $p$, Text embedding $C$, Label $y$}
\KwOut{Adversarially perturbed image $x_{\text{adv}}$}

\For{$\text{batch} \in \text{TrainLoader}$}{
    \For{$i \in \text{range}(\text{num\_iterations})$}{
        $t \leftarrow \text{randint}(0, 50)$\;
        $z \leftarrow \text{vae.encode}(x)$, $C \leftarrow \text{textencoder}(p)$\;
        $\epsilon \leftarrow \text{randn\_like}(z)$, $z_t \leftarrow \text{add\_noise}(z, \epsilon, t)$\;
        
        $\tilde{z} \leftarrow \sqrt{\frac{1}{\bar{\alpha}_t}} \left( z_t - \sqrt{1-\bar{\alpha}_t} \cdot \epsilon_\theta( z_t , t, C) \right)$\;
        
        $x_{\text{adv}} \leftarrow \text{vae.decode}(\tilde{z})$\;
        $x_{\text{adv}} \leftarrow (x_{\text{adv}} + 1) / 2$\;
        $x_{\text{adv}}.\text{Normalize}(\text{mean}, \text{std})$\;
        
        $\tilde{y} \leftarrow f(x_{\text{adv}})$\;
        $Loss \leftarrow -10^{10} \times \text{CrossEntropyLoss}(\tilde{y}, y) + \text{mse}(\epsilon, \epsilon_\theta( z_t , C))$\;
        
        $Loss.\text{backward}()$\;
    }
    \Return{$x_{\text{adv}}$}\;
}
\label{algorithm}
\end{algorithm}
\textbf{Attacking policy setup.}
Given a target image $\vect{x}$ and a latent diffusion model parameterized by $\epsilon_\theta$, our goal is to optimize the tokens within the conditioning text encoder's vocabulary to best represent the visual content of the image while simultaneously crafting an adversarial attack on the classifier. To determine the optimal tokens $\mathbf{P}^{*}$ for generating an adversarial image, we formulate the problem with two components. First, we employ a cross-entropy loss to measure the difference between the classifier's predicted output and the actual image class. Second, we include a loss that captures the discrepancy between the predicted noise from the network and the noise originally added to the image under a specific prompt condition, thereby ensuring semantic similarity between the adversarial image and the original image.
Our objective is to recover the set of embeddings $\mathbf{C}^{*}$ that align our objective of generating an adversarial image. Note that we freeze the diffusion model parameters throughout this optimization except the text encoder part. As discussed previously, instead of optimizing over all diffusion time steps using a standard SGD optimizer, we focus on the most impactful time steps, restricting the range of $t$ to later timesteps ($t \leq 50$). Subsequently, the image that shows the highest similarity to the original input is selected as the adversarial output. The complete approach of our method is illustrated in \cref{algorithm}.


\section{Experiments}

This section starts by outlining the experimental setup, the comparison framework, and the evaluation metrics used. We then present both quantitative and qualitative comparisons against the current state-of-the-art methods, focusing on the stealthiness of the perturbations introduced and the realism of the generated examples. Finally, we conduct ablation and sensitivity analyses to evaluate the influence of key components and hyperparameters.

\subsection{Experimental settings}

\textbf{Dataset and Models.} We leverage three publicly available breast ultrasound datasets and experiment with three widely recognized classifier architectures—ResNet34 \cite{resnet}, SqueezeNet1.1 \cite{SqueezeNet}, and DenseNet121 \cite{densenet}—which are commonly used as baselines in medical imaging research \cite{kriti2020deep, abhisheka2023comprehensive}. Given the relatively small dataset sizes, we apply five-fold cross-validation across all classes to ensure robust evaluation. For consistency with the classifier architectures, all images are resized to $224 \times 224$ pixels. Additionally, we incorporate a pre-trained LDM from Hugging Face, specifically utilizing the Stable-Diffusion-v-1-4 variant \cite{compvis2023stable}.
\\
\textbf{BUSI (Breast Ultrasound Images).} Collected in 2018, this dataset comprises ultrasound scans from 600 female patients between the ages of 25 and 75, gathered for breast cancer detection. The dataset includes a total of 780 images, each with an average resolution of 500x500 pixels, categorized into three classes: normal (133 images), benign (437 images), and malignant (210 images) \cite{Dataset}.
\\
\textbf{BUS-BRA.} BUS-BRA \cite{BUS-BRA} dataset includes 1875 anonymized images from 1,064 female patients, acquired with four different ultrasound scanners as part of systematic studies conducted at the National Institute of Cancer in Rio de Janeiro, Brazil. This dataset includes biopsy-confirmed tumor cases, with 1268 benign and 607 malignant cases.
\\
\textbf{UDIAT.} Collected in 2012 at the UDIAT Diagnostic Centre of the Parc Taulí Corporation in Sabadell, Spain, this dataset \cite{Dataset_C} was gathered using a Siemens ACUSON Sequoia C512 system with a 17L5 HD linear array transducer operating at 8.5 MHz. It comprises 163 images from 163 female patients, with an average resolution of 760x570 pixels, categorized into 53 malignant and 110 benign lesions.
\\
\textbf{Hyperparameters.} In our experiments, we train the model for 500 iterations with a learning rate of \(5 \times 10^{-4}\) using the AdamW optimizer \cite{loshchilov2017decoupled}. For adversarial methods, we set the FGSM attack parameter to \(\epsilon = 0.05\). For the PGD attack, we configure \(\epsilon = 0.03\) with a step size of \(\alpha = 0.007\) over 200 iterations. We utilize Diff-PGD with the same setting as it used for ImageNet \cite{deng2009imagenet}.
We run our experiments on a single NVIDIA RTX V100 GPU with 32GB memory.

\begin{table}[h]
\centering
\begin{tabular}{lccccc}
\toprule
\textbf{Attacker} & \textbf{Success Rate} & \textbf{LPIPS} & \textbf{SSIM} & \textbf{FID} \\
\midrule

\multicolumn{5}{c}{\textbf{DenseNet121}} \\
FGSM & 0.88 & 0.40 & 0.81 & 123.51 \\
PGD & 0.57 & 0.29 & 0.45  & 378.62 \\
Diff-PGD & \textbf{1.0} & 0.30  & \textbf{0.87} & 111.03 \\
P2P (Ours) & 0.98  &\textbf{0.13}  & 0.85 & \textbf{45.84} \\
\midrule
\multicolumn{5}{c}{\textbf{ResNet34}} \\
FGSM & 0.96  & 0.41 & 0.81 & 131.62 \\
PGD & 0.55 & 0.25 & 0.37  & 332.01 \\
Diff-PGD & \textbf{1.0} & 0.31 &  \textbf{0.84}& 117.49 \\
P2P (Ours) & 0.97  & \textbf{0.12} & 0.81 & \textbf{43.03} \\
\midrule
\multicolumn{5}{c}{\textbf{SqueezeNet1.1}} \\
FGSM & 0.49 & 0.16 & 0.40 & 118.03 \\
PGD & 0.33 & 0.20 & 0.30 & 250.38 \\
Diff-PGD & 0.74 & 0.14 & 0.56 & 79.51 \\
P2P (Ours) &\textbf{0.96} & \textbf{0.09}  & \textbf{0.63} & \textbf{47.64} \\
\bottomrule
\end{tabular}
\caption{Evaluation of adversarial attacks on different attack models for BUSI dataset \cite{Dataset} with 3 classifiers. LPIPS, SSIM, and FID are reported on successful attack examples.}
\label{table_a}

\end{table}
\begin{figure*}[!h]
    \centering
\includegraphics[width=0.92\textwidth]{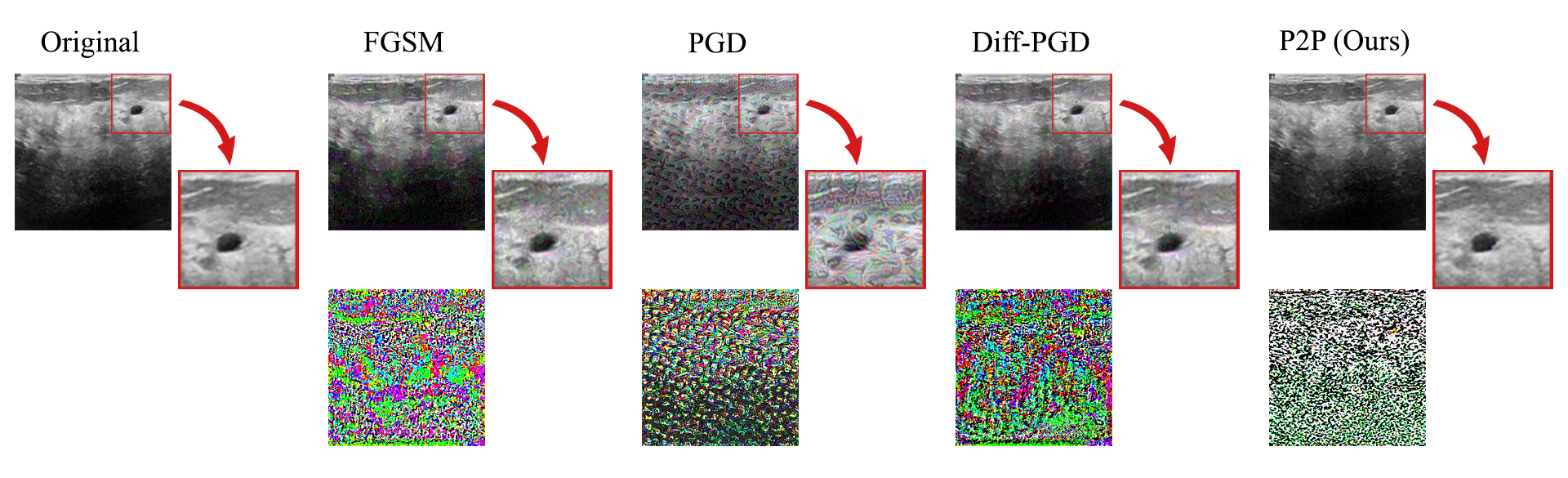}
    \vspace{-0.75em}
    \caption{Visual comparison of different attack methods on a benign image from the BUSI dataset, using DenseNet121 as the classifier. The second row displays the perturbations, calculated as the difference between the original image and the attacked example.} 
    \label{fig:attack-vis-comparison}
    \vspace{-1em}
\end{figure*}

\begin{figure*}[htb]
    \centering
\includegraphics[width=0.92\textwidth]{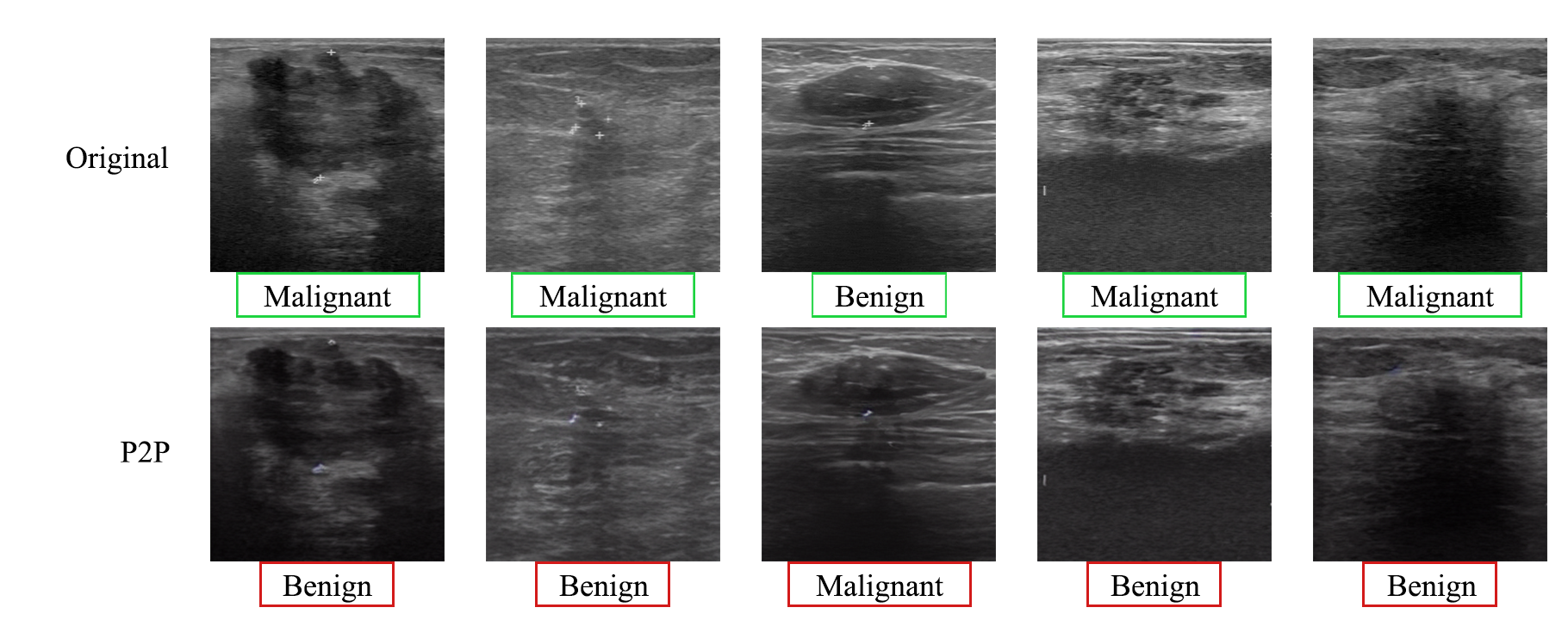}
    \vspace{-0.75em}
    \caption{Comparison of original and P2P-attacked ultrasound images from BUS-BRA Dataset, using DenseNet121 as the classifier. The top row shows the original images with their diagnostic labels, while the bottom row displays the same images after applying the P2P attack. Green boxes indicate the true labels, while red boxes show the labels predicted by the classifier after the attack.} 
    \label{fig:attack-examples}
    \vspace{-1em}
\end{figure*}

\subsection{Evaluation of Adversarial Examples} 
To evaluate the quality of adversarial examples generated by our proposed P2P method, we benchmark it against the leading Diffusion-Based Projected Gradient Descent (Diff-PGD) attack and conventional techniques like FGSM and PGD. We assess each attack's performance across metrics including Success Rate, LPIPS, SSIM, and FID, enabling a comprehensive analysis of both adversarial efficacy and the perceptual integrity of the generated adversarial samples.
\\
In our setting, each attack baseline is applied to validation set images that were initially classified correctly, with adversarial examples retained only when successful in misleading the classifiers. Accordingly, all metrics presented in \cref{table_a,table_b,table_c} reflect measurements derived exclusively from these modified images.
\\
Across all datasets and classifier architectures, our P2P approach consistently attains high Success Rates \cite{brown2017adversarial} comparable to those achieved by Diff-PGD, while surpassing baseline methods in Fréchet Inception Distance (FID) \cite{heusel2017gans}, reflecting the imperceptibility of the crafted adversarial examples, and in Learned Perceptual Image Patch Similarity (LPIPS) \cite{zhang2018unreasonable}, indicating reduced distortion in our attacks. These results highlight the superior perceptual quality of adversarial examples generated by our method.
\\
\begin{table}[h]
\centering
\begin{tabular}{lccccc}
\toprule
\textbf{Attacker} & \textbf{Success Rate} & \textbf{LPIPS} & \textbf{SSIM} & \textbf{FID} \\
\midrule

\multicolumn{5}{c}{\textbf{DenseNet121}} \\
FGSM &0.93  &0.40  &0.77  &112.11 \\
PGD &0.43  &0.19  & 0.56  &213.65  \\
Diff-PGD & \textbf{1.0} &0.29    &\textbf{0.82} &90.5  \\
P2P (Ours) &0.94  &\textbf{0.12}  &0.78  &\textbf{38.00} \\
\midrule
\multicolumn{5}{c}{\textbf{ResNet34}} \\
FGSM &0.81  &0.35  &0.66  &133.17 \\
PGD & 0.31 & 0.12 &  0.24 & 158.24 \\
Diff-PGD & \textbf{1.0}& 0.29  &\textbf{0.78} & 100.2 \\
P2P (Ours) &0.93  &\textbf{0.11}  &0.72  &\textbf{44.09} \\
\midrule
\multicolumn{5}{c}{\textbf{SqueezeNet1.1}} \\
FGSM &0.69  &0.16  &0.77  &120.14 \\
PGD & 0.43 &0.26  & 0.40  & 292.99 \\
Diff-PGD &\textbf{0.75} & 0.12  &0.47 & 89.47 \\
P2P (Ours) &0.74  &\textbf{0.08}  &\textbf{0.49}  &\textbf{58.60} \\

\bottomrule
\end{tabular}
\caption{ Evaluation of adversarial attacks on different attack models for BUS-BRA dataset \cite{BUS-BRA} with 3 classifiers. LPIPS, SSIM, and FID are reported on successful attack examples.
}
\label{table_b}
\end{table}
In terms of Structural Similarity Index Measure (SSIM), which evaluates the structural alignment between adversarial and original images, our method achieves competitive performance. Although Diff-PGD occasionally marginally outperforms P2P in SSIM for specific models, P2P consistently maintains strong SSIM values across all datasets, as shown in \cref{table_a,table_b,table_c}. This reflects high visual similarity between the generated adversarial examples \( x_{\text{adv}} \) and the original clean images \( x \), capturing low perceived visual distortion by preserving luminance, contrast, and structural details. The combination of high SSIM with low LPIPS and FID scores underscores P2P's ability to preserve structural consistency while producing perceptually aligned adversarial examples.
\\
For the DenseNet121 and ResNet34 classifiers, P2P achieves a strong balance of high Success Rate, low LPIPS, and favorable SSIM and FID scores, surpassing traditional methods. With SqueezeNet1.1, P2P exhibits particularly robust performance on perceptual metrics, reaching the lowest LPIPS and FID values across all datasets, as shown in  \cref{table_a,table_b,table_c}, highlighting its ability to generate realistic and structurally consistent adversarial images.
\\
Overall, P2P demonstrates competitive Success Rates while consistently producing high-quality adversarial examples with enhanced visual fidelity and structural similarity. The results underscore P2P’s effectiveness as a balanced approach for crafting effective and perceptually similar adversarial images across various datasets and classifiers. We will further elaborate on these results qualitatively in the subsequent section.

\begin{table}[h]
\centering
\begin{tabular}{lccccc}
\toprule
\textbf{Attacker} & \textbf{Success Rate} & \textbf{LPIPS} & \textbf{SSIM} & \textbf{FID} \\
\midrule

\multicolumn{5}{c}{\textbf{DenseNet121}} \\
FGSM &0.97 &0.37 &0.77 & 103.07 \\
PGD & 0.17 &0.07  & 0.11  & 147.84 \\
Diff-PGD &\textbf{1.0} &  0.27 &\textbf{0.80} &  81.31 \\
P2P (Ours) & 0.86 &\textbf{0.12}  &0.62  &\textbf{27.18} \\
\midrule
\multicolumn{5}{c}{\textbf{ResNet34}} \\
FGSM &0.98 &0.41 &0.75 & 103.68\\
PGD &0.23  &0.10  & 0.19  & 135.95 \\
Diff-PGD &\textbf{1.0} &0.31   &\textbf{0.77} & 80.89 \\
P2P (Ours) &0.97  &\textbf{0.15}  &0.74  &\textbf{20.3} \\
\midrule
\multicolumn{5}{c}{\textbf{SqueezeNet1.1}} \\
FGSM &0.59 &0.20 &0.42 & 72.51\\
PGD &0.16  &0.14  &0.19   & 292.21  \\
Diff-PGD &\textbf{0.77} &0.16 &\textbf{0.54}   & 32.47 \\
P2P (Ours) &0.76  & \textbf{0.10} &0.52  &\textbf{23.50} \\

\bottomrule
\end{tabular}
\caption{Evaluation of adversarial attacks on different attack models for UDIAT dataset \cite{Dataset_C} with 3 classifiers. LPIPS, SSIM, and FID are reported on successful attack examples.
}
\label{table_c}
\end{table}

\subsection{Qualitative Results of Adversarial Examples}
\begin{figure}[!htb]
    \centering
    \resizebox{0.46\textwidth}{!}{%
    \begin{tabular}{@{}cc@{}}
        \begin{minipage}{0.5\textwidth}
            \centering
            \includegraphics[width=1.0\textwidth]{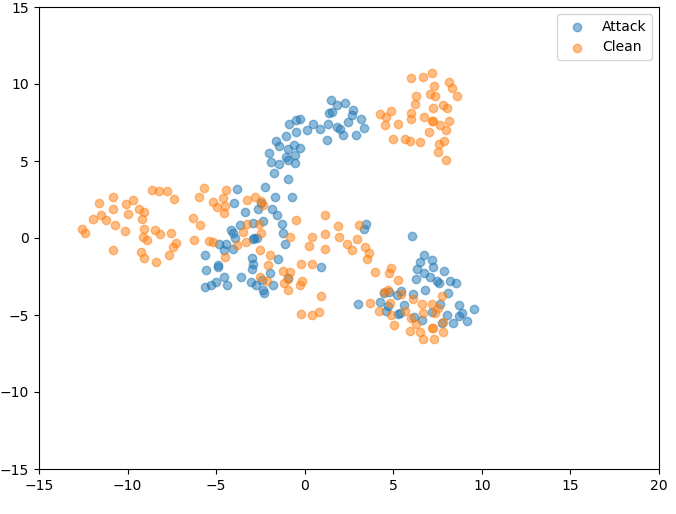}
            {\LARGE{ a) FGSM}}
        \end{minipage} &
        \begin{minipage}{0.5\textwidth}
            \centering
            \includegraphics[width=1.0\textwidth]{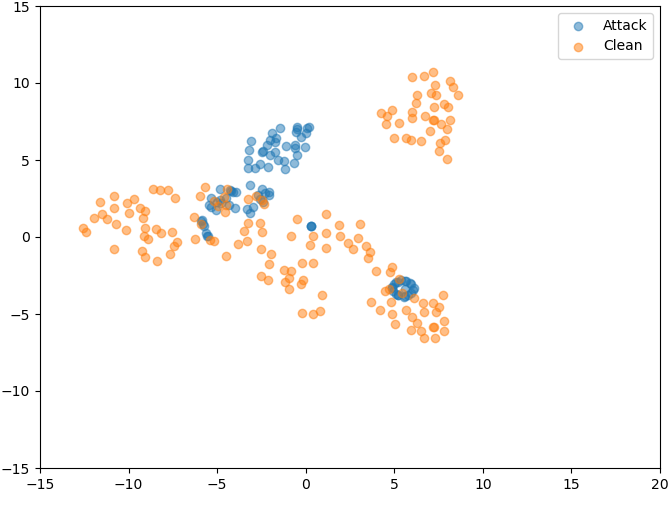}
            {\LARGE{ b) PGD}}
        \end{minipage} \\
        \begin{minipage}{0.5\textwidth}
            \centering
            \includegraphics[width=1.0\textwidth]{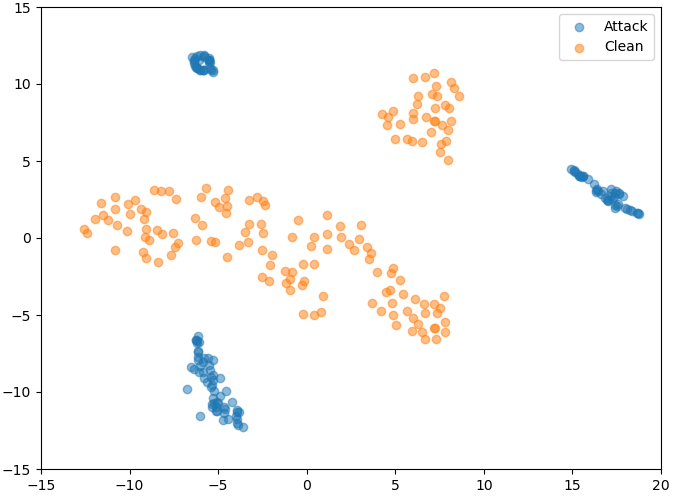}
            {\LARGE{ c) Diff-PGD}}
        \end{minipage} &
        \begin{minipage}{0.5\textwidth}
            \centering
            \includegraphics[width=1.0\textwidth]{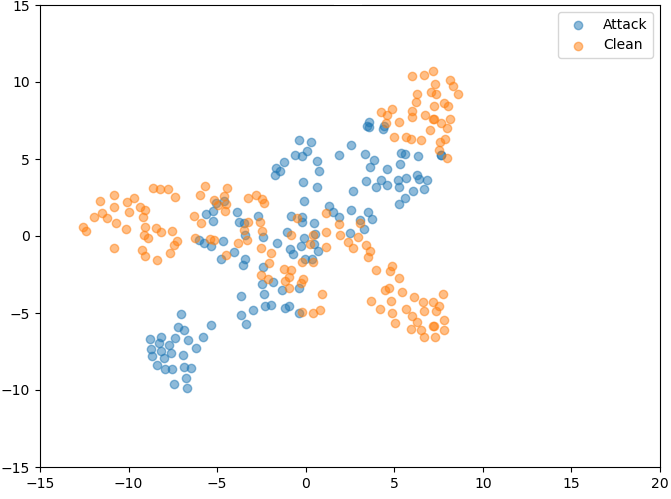}
            {\LARGE{ d) P2P (Ours)}}
        \end{minipage}
    \end{tabular}}
    \caption{t-SNE visualization of last-layer ResNet34 features on the BUSI dataset for FGSM, PGD, Diff-PGD, and P2P (Ours). Clean examples are shown in blue, and adversarial examples in orange.}
    \label{fig:four_figures}
\end{figure}

\begin{figure}[!htb]
    \centering
    \resizebox{0.46\textwidth}{!}{%
    \begin{tabular}{ccc}
        \begin{minipage}{0.3\textwidth}
            \centering
            \includegraphics[width=1.0\textwidth]{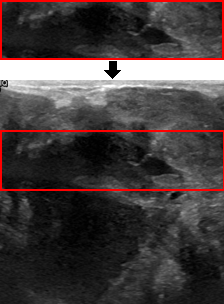}
            \\[5pt] 
            {\LARGE{ a) Original}} 
        \end{minipage} &
        \begin{minipage}{0.3\textwidth}
            \centering
            \includegraphics[width=1.0\textwidth]{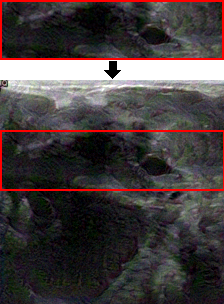}
            \\[5pt]
            {\LARGE{ b) Diff-PGD}} 
        \end{minipage} &
        \begin{minipage}{0.3\textwidth}
            \centering
            \includegraphics[width=1.0\textwidth]{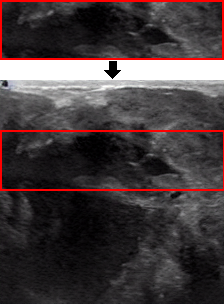}
            \\[5pt]
            { \LARGE{ c) P2P}} 
        \end{minipage}
    \end{tabular}}
    \caption{Attack examples of a malignant image of Dataset BUSI from Diff-PGD and P2P with ResNet34 as a classifier.}
    \label{fig:examples_train}
\end{figure}

\Cref{fig:attack-vis-comparison} presents visual comparisons of images and adversarial perturbations generated by different attack methods. Notably, our method (P2P) produces attack examples that more closely follow the distribution of the original medical images, resulting in a more natural and less detectable alteration. In contrast, other attack methods, such as FGSM, PGD, and Diff-PGD, introduce high-frequency noise with a distinct "textured" appearance. This textured noise can make these adversarial examples appear less natural, potentially revealing them as manipulated images, exhibit noisy artifacts. Besides, \cref{fig:attack-examples} shows that the P2P attack successfully changes the diagnostic labels of ultrasound images with minimal alteration to the images' semantic appearance, demonstrating label vulnerability despite preserved image semantics. Moreover, the t-SNE \cite{van2008visualizing} visualization of clean and attacked image features further confirms the effect of our prompt learning method. \Cref{fig:four_figures} compares four attack methods—FGSM, PGD, Diff-PGD, and P2P (Ours)—in terms of how well they blend attack samples with clean data clusters. Diff-PGD shows limited effectiveness, with noticeable separations between attack and clean samples. FGSM and PGD improve blending but still leaves isolated clusters of attack points. P2P achieves the best integration, with attack samples closely interwoven with clean data, indicating superior performance in making adversarial examples indistinguishable from clean data.
\\
To gain further insights, we examined attacked examples from the training set, focusing on images that the classifier identifies with high confidence. \Cref{fig:examples_train} presents malignant cases from the BUSI training set alongside their attacked versions generated by Diff-PGD and our method, P2P. As illustrated, the Diff-PGD attack introduces a noticeable noise pattern, while our method produces images that closely resemble the originals. Interestingly, additional examples reveal that Diff-PGD’s image quality deteriorates as the classifier’s confidence increases. While attack experiments are generally performed on validation sets, we analyzed training set images here to understand the performance of different attack models in varied scenarios.

\captionsetup[sub]{font=large}

\begin{table*}[!htb]
    \centering
    \resizebox{0.92\textwidth}{!}{%
    \begin{tabular}{@{}lccccccc@{}}
        \toprule
        \textbf{Component} & \textbf{Loss Function} & \textbf{Time Step}  & \textbf{Success Rate} & \textbf{LPIPS} & \textbf{SSIM} & \textbf{FID} & \textbf{Time (second)} \\ 
        \midrule
        Baseline &  &    & 0.97  & {0.12} & 0.81 & {43.03} & 178\\ 
        W/O MSE & \checkmark &  &   0.96 & 0.12 & 0.80 & 43.58 &178\\ 
        T=20 &  & \checkmark &   0.83 & 0.10 & 0.71 & 49.97 &  98 \\ 
        T=100 &  & \checkmark   &0.97  &0.12  &0.81  & 38.06 & 280 \\ 
        \bottomrule
    \end{tabular}%
    }
    \caption{Comparison of the ablation study on different components of the P2P pipeline. The baseline configuration uses T=50 with MSE in the loss function. In each row, only one component is modified from the baseline. 'Time' indicates the duration of the generation process for the attack per image. }\label{tab:ablation_comparison}
\end{table*}

\begin{figure}[!htb]
    \centering
    \resizebox{0.46\textwidth}{!}{%
    \begin{tabular}{ccc}
        \begin{minipage}{0.3\textwidth}
            \centering
            \includegraphics[width=1.0\textwidth]{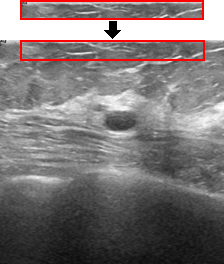}
            \\[5pt] 
            {\LARGE{ a) Original}} 
        \end{minipage} &
        \begin{minipage}{0.3\textwidth}
            \centering
            \includegraphics[width=1.0\textwidth]{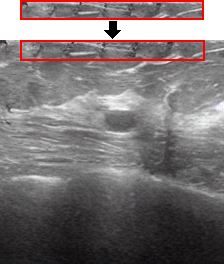}
            \\[5pt]
            {\LARGE{ b) W/O MSE}} 
        \end{minipage} &
        \begin{minipage}{0.3\textwidth}
            \centering
            \includegraphics[width=1.0\textwidth]{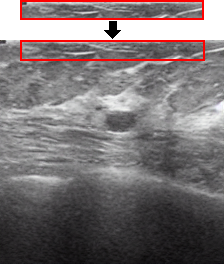}
            \\[5pt]
            {\LARGE{ c) MSE}} 
        \end{minipage}
    \end{tabular}}
    \caption{Impact of MSE loss in the P2P framework. Image (a) shows the clean reference image, image (b) shows the P2P result without MSE loss, and image (c) includes the MSE loss function.}
    \label{fig:mse}
\end{figure}

\subsection{Ablation Study}
This section disentangles the effects of design choices to assess their individual contributions to our approach's effectiveness. All ablation experiments are conducted on the BUSI dataset, using ResNet34 as the classifier, with results averaged over 5-fold cross-validation.
\\
\textbf{Loss Function.} Ablation study results on the effect of MSE loss function are reported in \cref{tab:ablation_comparison} (second row) and \cref{fig:mse}. Although using an MSE loss does not necessarily lead to better quantitative metrics, \cref{fig:mse} shows that it attempts to mitigate artifacts and retain the semantic integrity of the ultrasound image. Specifically, the middle image (without MSE) exhibits enhanced linear or streak-like artifacts, which may disrupt the homogeneity of tissue appearance (highlighted in the red rectangular region) and obscure finer structural details in the ultrasound image. Incorporating the MSE loss helps to reduce intensity fluctuations, creating a smoother and more consistent appearance, particularly beneficial in areas with subtle tissue boundaries.
\\
\textbf{Step Selection} To justify the choices for P2P with optimization for different timesteps, we show in \cref{tab:ablation_comparison} the
performance of our method with different settings. Using $20$ steps, the approach yields prompts that generate attack images with relatively low success, SSIM and LPIPS. However, when optimizing with more timesteps ($T=100$) even though, the FID of the generated attack images is better, other metrics are comparable to that of ($T=50$), but with longer time to generate attack images. Therefore, given the efficiency dilemma and the results presented, selecting $50$ steps appears to be a reasonable choice.
\\
\section{Conclusion}

In this paper, we designed an effective adversarial attack procedure called \textbf{P2P} for synthesizing adversarial breast ultrasound images using prompt learning in text-to-image diffusion models.
We showcase the effectiveness of our prompt-learning approach across various attack scenarios, especially valuable in data-limited medical applications. Our adversarial images surpass previous methods, appearing more natural and harder to detect. They also maintain lower perceptual differences while achieving greater structural similarity. P2P enables significantly easier attacks on classifiers within the medical domain, eliminating the need for pre-trained models in a specific domain. By facilitating the construction of effective textual examples, it enhances the ability to target and challenge classifiers directly.
\section*{Acknowledgments}
\small
We acknowledge the support of the Natural Sciences and Engineering Research Council of Canada (NSERC), [funding reference number RGPIN-2023-03575].
Cette recherche a été financée par le Conseil de recherches en sciences naturelles et en génie du Canada (CRSNG), [numéro de référence RGPIN-2023-03575].
{
    \small
    \bibliographystyle{ieeenat_fullname}
    \bibliography{main}
}


\end{document}